\documentclass{scrarticle}

\usepackage[english]{babel}
\usepackage{natbib}
\usepackage{CJKutf8}

\usepackage{subcaption}

\usepackage{tabularx}

\usepackage[letterpaper,top=2cm,bottom=2cm,left=3cm,right=3cm,marginparwidth=1.75cm]{geometry}

\usepackage{amsmath}
\usepackage{graphicx}
\usepackage[colorlinks=true, allcolors=blue]{hyperref}

\title{Is deeper always better? Replacing linear mappings with deep learning networks in the Discriminative Lexicon Model}
\author{Maria Heitmeier, Valeria Schmidt,\\Hendrik P.A. Lensch and R. Harald Baayen}
\date{}

\begin{document}
\maketitle

\begin{abstract}
Recently, deep learning models have increasingly been used in cognitive modelling of language. This study asks whether deep learning can help us to better understand the learning problem that needs to be solved by speakers, above and beyond linear methods. We utilise the Discriminative Lexicon Model introduced by Baayen and colleagues, which models comprehension and production with mappings between numeric form and meaning vectors. While so far, these mappings have been linear (Linear Discriminative Learning, LDL), in the present study we replace them with deep dense neural networks (Deep Discriminative Learning, DDL). We find that DDL affords more accurate mappings for large and diverse datasets from English and Dutch, but not necessarily for Estonian and Taiwan Mandarin. DDL outperforms LDL in particular for words with pseudo-morphological structure such as chol+er. Applied to average reaction times, we find that DDL is outperformed by frequency-informed linear mappings (FIL). However, DDL trained in a frequency-informed way (‘frequency-informed’ deep learning, FIDDL) substantially outperforms FIL. Finally, while linear mappings can very effectively be updated from trial-to-trial to model incremental lexical learning, deep mappings cannot do so as effectively. At present, both linear and deep mappings are informative for understanding language.
\end{abstract}

\section{Introduction}

Over the last decade, deep learning models\footnote{Throughout this paper, we use the term `deep learning' to refer to neural networks with at least one hidden layer (with a non-linear activation function). While it can be argued whether a neural network with a single hidden layer can indeed be called `deep', the definitions of deep learning models usually only require such models to have ``multiple (non-linear) processing layers'' \citep[e.g.][]{arnold2011introduction, lecun2015deep} and for reasons of simplicity we therefore include models with a single non-linear hidden layer in the definition. Crucially, and contrary to models without any hidden layers and no activation functions, such models are able to learn non-linear functions \citep[e.g.][]{webb1990optimised, widrow199030}.} have been remarkably successful, culminating in the most recent large language models \citep[e.g.][]{openai2023gpt4, anthropic2023claude2, team2023gemini}. Increasingly, deep learning has also made its way into cognitive modelling of language with studies exploring whether deep learning architectures can model aspects of language processing. Modelling efforts have been targeted at roughly four groups of models: predicting how humans form known and novel wordforms, such as inflections \citep[e.g.][]{rumelhart1986learning, cardillo2018deep, kirov2018recurrent, corkery2019we, mccurdy2020inflecting}, using activations in deep learning models to directly predict behavioural data or brain signals \citep[e.g.][]{bhatia2022transformer, salewski2024context, huang2024large, caucheteux2020language, schrimpf2021neural, hale2022neurocomputational, goldstein2022shared}, using deep learning components in theoretical models of lexical processing \citep[e.g.][]{seidenberg1989distributed, harm2004computing, magnuson2020earshot, scharenborg2008modelling}, and finally, leveraging the iterative learning ability of deep learning models to model language acquisition \citep[e.g.][]{plunkett1992symbol, li2007dynamic}.

While these efforts have often been quite successful, using deep learning comes at the cost of ease of interpretation \citep[but see e.g.][]{cammarata2020thread:, elhage2021mathematical, bricken2023monosemanticity, linzen2021syntactic}. Deep learning models are universal function approximators, implying that if powerful enough, they can approximate any continous function with arbitrary accuracy \citep{hornik1989multilayer}. For cognitive modelling, however, pure accuracy is not the end-goal: the question is whether increased precision afforded by deep learning models actually helps with predicting (cognitive) behaviour.

In the present paper, we address this question for the Discriminative Lexicon Model \citep[DLM,][]{Baayen2018a, baayen2019discriminative}, a model of (single) word comprehension and production. The model has been used to model the morphology of a wide range of languages \citep[e.g.][]{chuang2020estonian, nieder2021comprehension, Heitmeier:Chuang:Axen:Baayen:2022, chuang2021vector, vijver2022word} but has also been successful at predicting behavioral data such as reaction times and acoustic durations \citep[e.g.][]{gahl2024time, stein2021morpho, heitmeier2023trial, cassani2019nonwords, chuang2020processing, baayen2020modeling, Saito:2023}. Word forms are represented by high-dimensional binary vectors and semantics by high-dimensional real-valued vectors. In the DLM's original formulation, comprehension is modelled with a linear mapping from word forms to word meanings and vice versa for production. Linear mappings are highly interpretable and surprisingly powerful, but also inherently limited by their linearity. In the present study we use the properties of linear and deep learning to investigate whether computational modeling can help us to better understand the learning problem that needs to be solved by speakers. Since we investigate single-word learning rather than word sequences and we believe that in the context of cognitive modelling the analysability of models is very important, we focus on using simple dense neural networks rather than more complex deep learning architectures such as Recurrent Neural Networks (RNNs) or transformers. We are specifically interested in whether, and under what circumstances, deep learning generates insights above and beyond what more easily interpretable, linear models can tell us.

This paper is structured as follows: Section~\ref{sec:methods} briefly describes the DLM, the dense neural networks (termed Deep Discriminative Learning, DDL) we used to replace the linear mappings and how they were trained. Section~\ref{sec:internal} shows that DDL generally affords higher mapping accuracy, particularly on the training data, and explores for which words DDL outperforms LDL. Section~\ref{sec:external} investigates DDL's ability to predict lexical decision reaction times, both averaged across individuals and on a trial-by-trial basis. Finally, Section~\ref{sec:discussion} discusses the results and concludes the paper.

\section{Methods}\label{sec:methods}

\subsection{Data}\label{sec:data}

In the present study we used four datasets\footnote{Data, code and analysis scripts can be found in the Supplementary Materials at \url{https://osf.io/tyfdr/}).}:\\
\textbf{British Lexicon Project (BLP)}: Large-scale British English visual lexical decision database \citep{keuleers2012british}. The final dataset contained 28,482 word forms. We modelled this dataset with trigrams and word embeddings grounded in vision from \citet{shahmohammadi2021learning}.\\
\textbf{Dutch Lexicon Project (DLP)}: Large-scale Dutch visual lexical decision database \citep{keuleers2010practice}. The final dataset contained 13,675 word forms. We modelled this dataset with trigrams and fasttext embeddings \citep{grave2018learning}.\\
\textbf{Estonian}: This dataset was first introduced in \citet{chuang2020estonian}. The final dataset contained 4,093 wordforms of
230 lemmas, thus on average 17.8 wordforms per lemma (range from 2 to 28). We modelled this dataset with trigrams and fasttext embeddings \citep{grave2018learning}.\\
\textbf{Taiwan Mandarin}: This dataset was first introduced in \citet{chuang2021discriminative} based on \citet{fon2004preliminary}. It contained 7,349 words with one to five syllables. We used a phone-based representation where vowels are combined with a number indicating lexical tone (e.g. \begin{CJK*}{UTF8}{gbsn} 一下子\end{CJK*}, eng. `abruptly', is represented as \texttt{i1.x.ia4.z.ii5}) and fasttext embeddings \citep{grave2018learning}.

The four datasets pose different challenges: English and Dutch have limited inflectional morphology, therefore the challenge of the BLP and DLP datasets lies in modelling a large number of different lemmas. The dataset for Estonian, with its rich inflectional system, challenges our models to accurately map many different forms for only a few different lemmas. Finally, the challenge of the Mandarin dataset lies in modelling both segmental (phone) and suprasegemental (tone) information.

\subsection{Models}

In the Discriminative Lexicon Model (DLM) comprehension is modelled as a mapping from a word's form to its meaning, and production as a mapping from meaning to form. The simplest way to represent wordforms is by first splitting wordforms into letter/phone n-grams and then marking the presence and absence of n-grams in high-dimensional binary vectors. The form vectors of all wordforms in the lexicon are stored in a form matrix $\mathbf{C}$. Word meanings are likewise represented by high-dimensional vectors, usually taken from pre-trained word embedding spaces such as word2vec or fasttext \citep{mikolov2013distributed, grave2018learning}. Details on which embedding vectors were used for which dataset can be found in Section~\ref{sec:data}. The embedding vectors are then likewise stored in a semantic matrix $\mathbf{S}$. In Linear Discriminative Learning (LDL), comprehension is then modelled via a linear mapping $\mathbf{F}$ such that $$\hat{\mathbf{S}} = \mathbf{CF}$$ and production via $\mathbf{G}$
$$\hat{\mathbf{C}} = \mathbf{SG}$$
with $\hat{\mathbf{S}}$ and $\hat{\mathbf{C}}$ indicating predicted $\mathbf{S}$ and $\mathbf{C}$ matrices respectively, borrowing notation from statistics. There are several methods available for calculating $\mathbf{F}$ and $\mathbf{G}$, such as standard least squares regression which optimises across all word forms equally, or ``frequency-informed learning'' which uses a form of weighted regression to inform the mappings about word forms' frequencies \citep{Heitmeier:Chuang:Axen:Baayen:2022}. We used various methods throughout this study and give more detail in the relevant sections.

Once $\mathbf{F}$ and $\mathbf{G}$ are in place, they can be used to predict semantic and form vectors for seen or unseen word forms $w$ via $$\hat{\mathbf{s}}_w = \mathbf{c}_w\mathbf{F}$$ $$\hat{\mathbf{c}}_w = \mathbf{s}_w\mathbf{G}$$ where $\mathbf{c}_w$ and $\mathbf{s}_w$ are the form and meaning vectors of $w$ and $\hat{\mathbf{s}}_w$ and $\hat{\mathbf{c}}_w$ the model's predicted vectors. A thorough introduction to LDL can be found in \citet{heitmeier2023linear}.

In Deep Discriminative Learning (DDL) we replace $\mathbf{F}$ and $\mathbf{G}$ with dense neural networks\footnote{Note that we also experimented with LSTMs \citep{hochreiter1997long} and transformer architectures \citep{vaswani2017attention} but ultimately found that for the problem at hand simple dense neural networks provided the best generalisation accuracy while also being directly comparable to the linear setup.}: \begin{align*}\hat{\mathbf{S}} &= f(\mathbf{C})\\
\hat{\mathbf{C}} &= g(\mathbf{S}).\end{align*}
\noindent Since multi-layer neural networks can in principle approximate any mathematical function \citep[with some limitations,][]{hornik1989multilayer}, we denote them with $f$ and $g$.

\begin{figure}[!ht]
    \begin{subfigure}[T]{\textwidth}
    \centering
    \includegraphics[width=\textwidth]{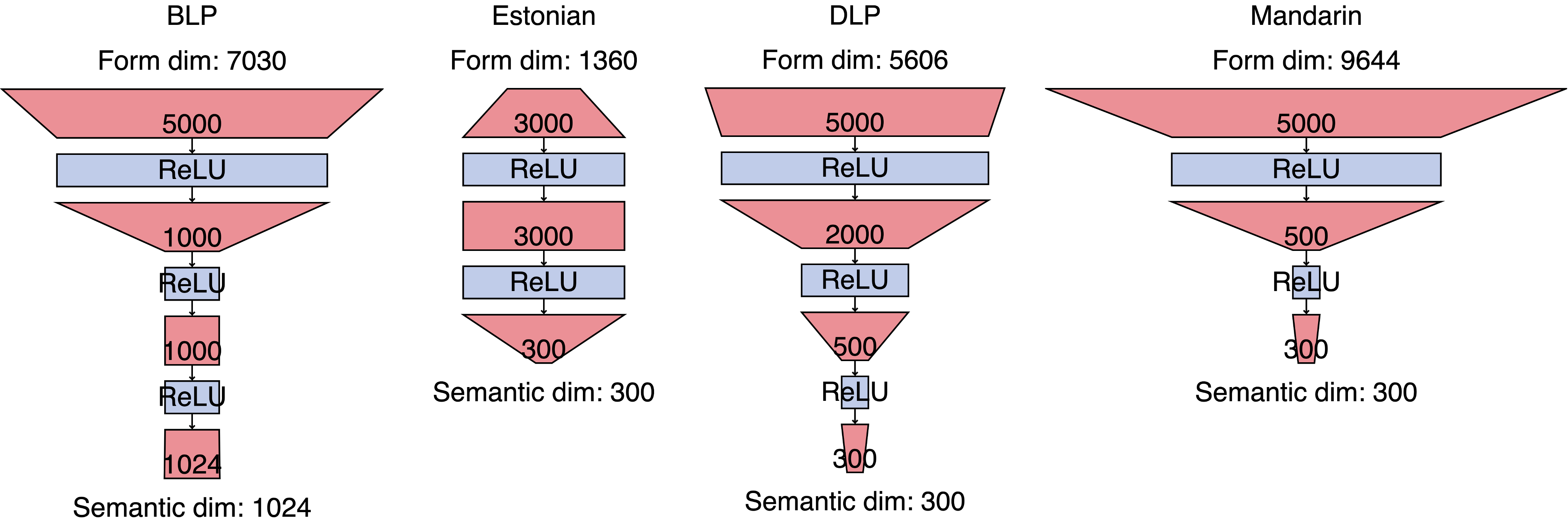}
       \caption{Comprehension}
       \label{fig:comp}
    \end{subfigure}\hfill
    \begin{subtable}[T]{\textwidth}
    \centering
    \includegraphics[width=\textwidth]{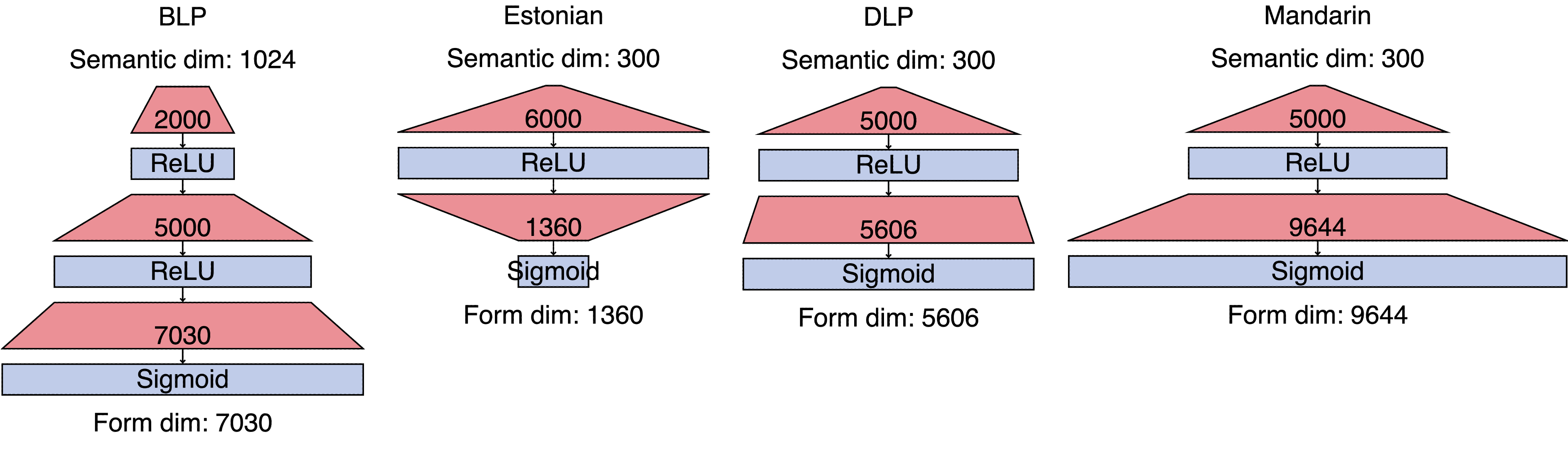}
        \caption{Production}
        \label{fig:prod}
     \end{subtable}
     \caption{DDL model architectures. The red boxes represent dense layers (matrix multiplication plus bias) with the output dimension printed inside, blue boxes represent non-linearities. Box widths reflect the input and output dimensions of the layers. Figure from \citet{heitmeier2024}.}
     \label{fig:model_architectures}
\end{figure}

Please refer to Figure~\ref{fig:model_architectures} for an overview of the number of hidden layers and dimensions for each model in this paper (one model per dataset and comprehension/production). We used Rectified Linear Units (ReLUs)\footnote{Due to space limitations, we refrain here from defining terms commonly used in machine learning literature. A more accessible introduction to Deep Discriminative Learning including term definitions can be found in Chapter 11 of \citet{heitmeier2023linear}.} as activation function throughout. The comprehension models were trained with a mean squared error loss. For production models, we opted to regard the prediction of which n-grams are part of a wordform as a multi-class multi-label classification problem where all n-grams are ``classes'' and multiple n-grams can be selected as ``labels''. As commonly done in multi-class multi-label classification models, we therefore added a sigmoid function after the final layer of the production models to restrict the final predicted values between 0 and 1 (approximating probabilities) and trained them with a binary cross-entropy loss. Note that adding a sigmoid + binary cross-entropy loss to the model in this way does not add any additional trainable parameters.
We used a batchsize of 512, except for the Estonian dataset where we used a batchsize of 256. Unless indicated otherwise, we used the Adam optimizer \citep{kingma2014adam} with a learning rate of 0.001. Further justification for these modelling choices can be found in \citet{heitmeier2024}. Other training details depend on the experiment and are laid out in the sections below.

We note here that we do not regard the different architectures across the different datasets and tasks as being informative about the cognition of speakers of the respective languages. Instead, we think that the architectures are determined by the datasets. The datasets vary considerably in size which influences the networks' required sizes. The languages' structures presumably also influence the architectures, but even regarding this point, we are wary of making any claims about speakers' cognition given that the kind of neural networks we are using are not particularly biologically plausible and how words were selected varies considerably across datasets. Instead, we view the presented architectures as the optimal models for solving the specific mapping problems for the specific datasets (within the scope of hyperparameters tested). As such, we are interested in what can theoretically be learned by a statistical system and testing whether what is learned by such a system allows insights about behaviour as well.

\section{Does DDL improve mapping accuracy?}\label{sec:internal}

\begin{figure}[!ht]
    \begin{subfigure}{\textwidth}
        \centering
    \includegraphics[width=\linewidth]{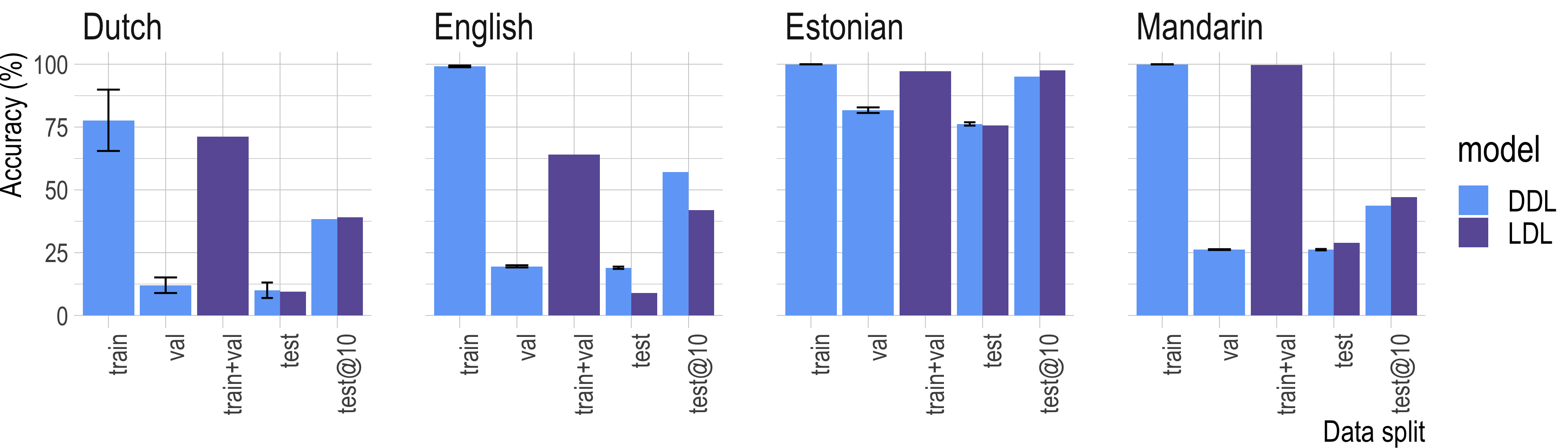}
    \caption{Comprehension}
    \end{subfigure}
    \begin{subfigure}{\textwidth}
        \centering
    \includegraphics[width=\linewidth]{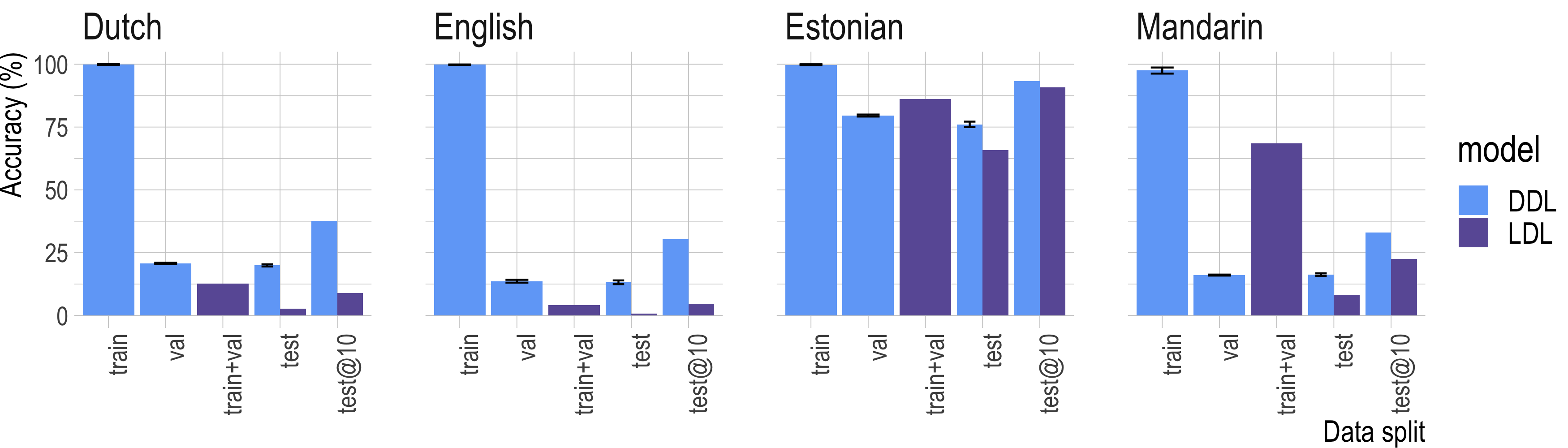}
    \caption{Production}
    \end{subfigure}
    \caption{Correlation accuracy for DDL and LDL models for a British English, Dutch, Estonian and Taiwan Mandarin dataset. For DDL models, accuracies are averaged across 10 models, the error bar shows the standard deviation. DDL clearly outperforms LDL for Dutch and English, and less clearly for Estonian and Mandarin. Test@10 indicates the accuracy@10 on the test data for the best model.}
    \label{fig:comp_prod_internal}
\end{figure}

First, we need to verify that DDL does indeed improve accuracy of the mappings compared to LDL. Here, we focus both on performance on seen data as well as unseen data. We split the data into an approximate 80/10/10 split for training/validation/test data. The data was split in such a way that all letter n-grams in the validation and test data also occurred in the training data. We trained models for all four languages for a maximum of 500 epochs, with early stopping when validation accuracy did not increase for 20 epochs for the comprehension and 50 epochs for the production models. As a baseline, we also trained LDL models on the joint training and validation data (since LDL does not require a validation set)\footnote{Even though DDL is not directly trained on the validation data, it is nevertheless used in training to select the epoch after which to stop training when performance on the validation data is best. Thus, even though the validation data is not directly used to calculate the loss and update the model's weights, it nevertheless influences the trained model. Due to this crucial role in the \textit{training} of the DDL models,
it is important to train the linear models on both the training and validation data.}. We evaluated all models using correlation accuracy: if the predicted vector was more correlated with its target vector than with all vectors in the remaining data, it was counted as correct. We also computed accuracy@10 for the best models for each datasets. For accuracy@10, a mapping is counted as correct, if the target vector is among the 10 most correlated vectors. The results can be found in Figure~\ref{fig:comp_prod_internal}.

It is clear that DDL outperforms LDL for comprehension only for the English data where the challenge lies in mapping a diverse set of meanings. For Estonian and Mandarin on the other hand, LDL already performs surprisingly well and DDL only improves on LDL for production. However, DDL outperforms LDL for all datasets in production and the general increase in training accuracy across all four languages indicates that DDL does indeed afford higher mapping precision.

\begin{figure}[!ht]
    \centering
    \includegraphics[width=\linewidth]{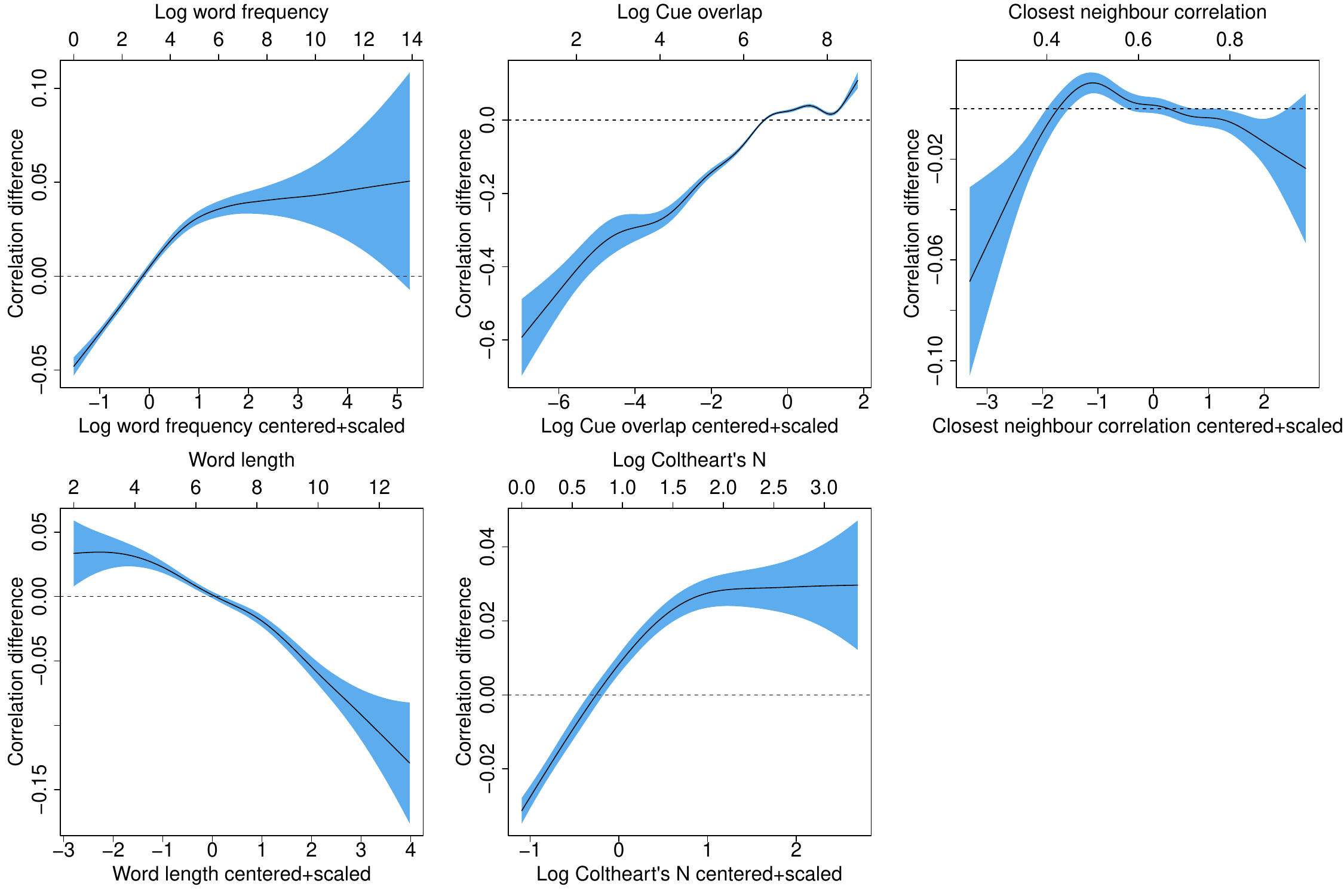}
    \caption{Measures of between-word similarity predicting the difference between target correlation in DDL and LDL (target correlation DDL - target correlation LDL). For higher values of correlation difference, DDL outperforms LDL. The lower x-axes show the centered and scaled values as entered into the GAMs; the upper x-axes show the original values for reference. We find that in general, DDL outperforms LDL for words that are similar to other words (i.e. words with higher cue overlap, a dense orthographic neighbourhood, shorter words and more frequent words). Figure adapted from \citet{heitmeier2024}.}
    \label{fig:BLP_diff_comp}
\end{figure}

Next, we investigated for which words DDL outperforms LDL. We hypothesised that DDL would perform better than LDL for words which are harder to distinguish from other words. These are words with n-grams that are shared with many other words, shorter words and words in a dense semantic neighbourhood. We used these predictors (plus word frequency as a control) to predict the difference in correlation between predicted and target vectors in LDL and DDL using a Generalised Additive Model (GAM, \citealp{wood2011gam}; more details on how we operationalised these concepts can be found in \citealp{heitmeier2024}). Figure~\ref{fig:BLP_diff_comp} shows that DDL indeed outperforms LDL for words with higher cue overlap (i.e. a word's cues occur in many other words), shorter words and words in a denser orthographic neighbourhood \citep[higher Coltheart's N,][]{coltheart1977access}. Moreover, it performs better on words with higher frequency. While semantic density (measured as the correlation between the target semantic vector and its closest semantic neighbour) seems to have a U-shaped effect, this is due to collinearity with the other predictors \citep{friedman2005graphical}: in a GAM with only semantic density as a predictor, the measure shows the expected effect of an advantage for DDL for words in denser semantic neighbourhood.

We also qualitatively examined words that showed particularly stark differences in correlation, favoring either LDL or DDL. The results can be found in Table~\ref{tab:corr_diff_words}. LDL outperforms DDL if words contain unique cues such as \textit{px} or \textit{oz}. DDL on the other hand shows a clear advantage for words which contain suffixes expressing a range of different meanings such as \textit{barriers} or \textit{choler} which all contain trigrams which are often used in comparatives. Consequently, LDL places them in the semantic neighbourhood of other comparatives, while DDL maps them to their correct neighbourhood. This ability of DDL to work around the re-use of trigrams to express different meanings points to an interesting issue: language is full of re-use of sub-lexical features, such as the \textit{-s} ending, which can express plurality \citep[see also][]{shafaei2022semantic,plag2015homophony}, third person singular, or possessive or simply be part of a word’s stem, or the \textit{-er} ending, which can express varied meanings such as comparative (e.g., \textit{hungrier}), agent (e.g., \textit{teacher}), instrument (e.g., \textit{poker}), or something associated with an action (e.g., \textit{diner}) or can denote the origin of a person (e.g., \textit{Londoner}) \citep[p.89]{plag2003word}. Any system modelling morphology needs to be able to deal with this ``bricolage'' \citep{strauss1962savage}. Apparently, DDL is much more effective at doing so than LDL. This does not seem to be just an issue of precision: both Table~\ref{tab:corr_diff_words} and the accuracies@10 in Figure~\ref{fig:comp_prod_internal} suggest that for LDL, the correct target word does not even make it to the top 10 most correlated words. A replication of this analysis for Dutch can be found in the Supplementary Materials.

\begin{table}[!ht]
    \centering
    \scriptsize
    \begin{subtable}[h]{\textwidth}
        \centering
\begin{tabularx}{\textwidth}{l|X|X|r}
  \hline
  Word & Semantic neighbours LDL & Semantic neighbours DDL & Correlation difference\\
  \hline\hline
by & by, standby, bye, written, thereby, brigades, dismissed, copied, outplayed, routed & bypassed, policed, weeded, cleared, entrenched, enforced, watered, encroached, deterred, imposed & -0.51 \\
  px & px, width, height, border, size, square, margin, padding, serif, background & px, margin, apex, equals, border, serif, font, zero, cornered, width & -0.50 \\
  my & my, her, me, your, own, myself, his, friend, our, dearest & my, friend, dearest, dear, proudest, thankful, hearts, beloved, fathers, coolest & -0.44 \\
  if & if, unless, sure, else, whether, because, assume, not, does, should & if, unless, assume, sure, whereas, therefore, neither, likely, maybe, suppose & -0.40 \\
  oz & oz, ounce, ounces, bottle, quart, gallon, jar, servings, cologne, spray & oz, ounce, ounces, bottle, juice, gallon, shampoo, tumbler, cleanser, pint & -0.38 \\
   \hline
\end{tabularx}

       \caption{LDL outperforms DDL}
       \label{tab:ldl_over_ddl}
    \end{subtable}
    \begin{subtable}[h]{\textwidth}
        \centering
\begin{tabularx}{\textwidth}{l|X|X|r}
  \hline
 Word & Semantic neighbours LDL & Semantic neighbours DDL & Correlation difference\\
  \hline\hline
lungs & lunches, lunge, slogs, dragged, luncheons, lunges, lurches, clamoured, jaunts, dives & lungs, lung, chest, breath, windpipe, coughing, exhale, larynx, breathe, throat & 0.54 \\
  scarves & carves, carving, starves, carvers, carver, corpses, gouges, carved, sews, sculpts & scarves, scarfs, sweaters, stoles, blankets, hats, headbands, afghans, robes, capes & 0.55 \\
  leans & leaches, cleanses, scoundrels, deign, nazis, cleans, traitors, betters, purveys, kikes & leans, pulls, mutters, nods, pushes, straightens, crouches, grabs, squeezes, smirks & 0.56 \\
  barriers & livelier, chattier, sturdier, flashier, rowdier, trendier, sketchier, ruder, heartier, fleshier & barriers, barrier, impede, roadblocks, blocking, curbs, hurdle, shackles, hinder, doorways & 0.56 \\
  choler & chunkier, wiser, moister, heartier, blunter, fattier, gunsmith, braver, plainer, trimmer & choler, slackness, rancour, wherefore, quinsy, foulness, rapine, glumness, stoutness, largeness & 0.59 \\
   \hline
\end{tabularx}
        \caption{DDL outperforms LDL}
        \label{tab:ddl_over_ldl}
     \end{subtable}
    \caption{Words for which LDL outperforms DDL the most (a) and for which DDL outperforms LDL the most (b). The two ``semantic neighbours'' columns show the words of the 10 semantic vectors which are most correlated with the vectors predicted by LDL and DDL respectively. The rightmost column shows the difference in target correlation between LDL and DDL. LDL outperforms DDL for words with unique n-grams, while DDL outperforms LDL for words with endings which can express various meanings (e.g. the \textit{-er} ending). In the latter case, LDL tends to place words in the semantic neighbour of words with the same ending, while DDL puts them in the neighbourhood of words with similar meanings.}
    \label{tab:corr_diff_words}
\end{table}

\section{Does DDL improve prediction of behavioural data?}\label{sec:external}

Having ascertained that DDL does indeed generally improve mapping precision, particularly for the BLP and DLP, we were interested next in whether this also means that DDL outperforms LDL at predicting behavioural data. To this end we used visual lexical decision data from the BLP and DLP. In the following, we present results of predicting reaction times averaged across participants and predicting reaction times on a trial-by-trial basis following \citet{heitmeier2023trial}.

\subsection{Predicting average reaction times}

For both the DLP and BLP we trained five comprehension models:
\begin{itemize}
\setlength\itemsep{0em}
    \item Endstate-of-Learning (EL): LDL model optimised equally across all words in the DLP/BLP.
    \item Frequency-informed learning (FIL): LDL model trained in a frequency-informed way using a form of weighted regression \citep{Heitmeier:Chuang:Axen:Baayen:2022}.
    \item Deep Discriminative Learning (DDL): DDL model trained on the 90\% most frequent words in the DLP/BLP. Training was stopped after accuracy on the remaining 10\% of words did not improve for 20 epochs.
    \item Endstate-DDL (EDDL): DDL model trained on all words in the DLP/BLP for 2000 epochs to model a theoretical endstate of learning where all words are learned optimally.
    \item Frequency-informed DDL (FIDDL): DDL model trained on the full token distribution of words in the DLP/BLP, i.e. the model was trained on each word as many times as its frequency count.
\end{itemize}

\noindent
We used CELEX \citep{baayen1995celex} frequencies for both the BLP and DLP throughout. After training, we obtained the correlations between the predicted and the target semantic vectors for each model (``target correlation''). Reaction times were transformed as follows:

$$\text{RTinv} = \frac{-1000}{\text{RT}}$$

\noindent
We used the target correlation measure to predict RTinv in one GAM per discriminative learning model.
Target correlation was a highly significant predictor for reaction times across all models and both datasets ($p < 0.0001$; summary tables for the GAMs can be found in the Supplementary Materials). Figure~\ref{fig:dlp_blp_rts} shows the fit of the five models in terms of the Akaike Information Criterion \citep[AIC,][]{akaike1998information}\footnote{AIC measures model fit while punishing model complexity. Lower AIC values indicate better model fit.} for both the DLP and BLP. First, we see that for both datasets, frequency-informed methods (FIL, FIDDL) outperform their non-frequency-informed counterparts (EL, DDL, EDDL). Regarding the difference between DDL and LDL, DDL and EDDL outperform EL for the DLP, but not for the BLP. Interestingly, while the correlation accuracy for EDDL was clearly higher than for DDL (DLP: 100\% for EDDL compared to 93\% for DDL; BLP: 100\% compared to 89\%), when it comes to predicting reaction times, EDDL only shows a slight advantage over DDL for the BLP and performs worse for the DLP. This indicates that better mapping accuracy does not necessarily entail better performance at predicting behavioural data. FIDDL outperforms all other models. From this we can conclude that while DDL does not necessarily outperform LDL, FIDDL clearly outperforms LDL both without and with frequency information. We will return to both these points in the final discussion.

\begin{figure}
    \begin{subfigure}{.45\textwidth}
        \centering
        \includegraphics[width=\textwidth]{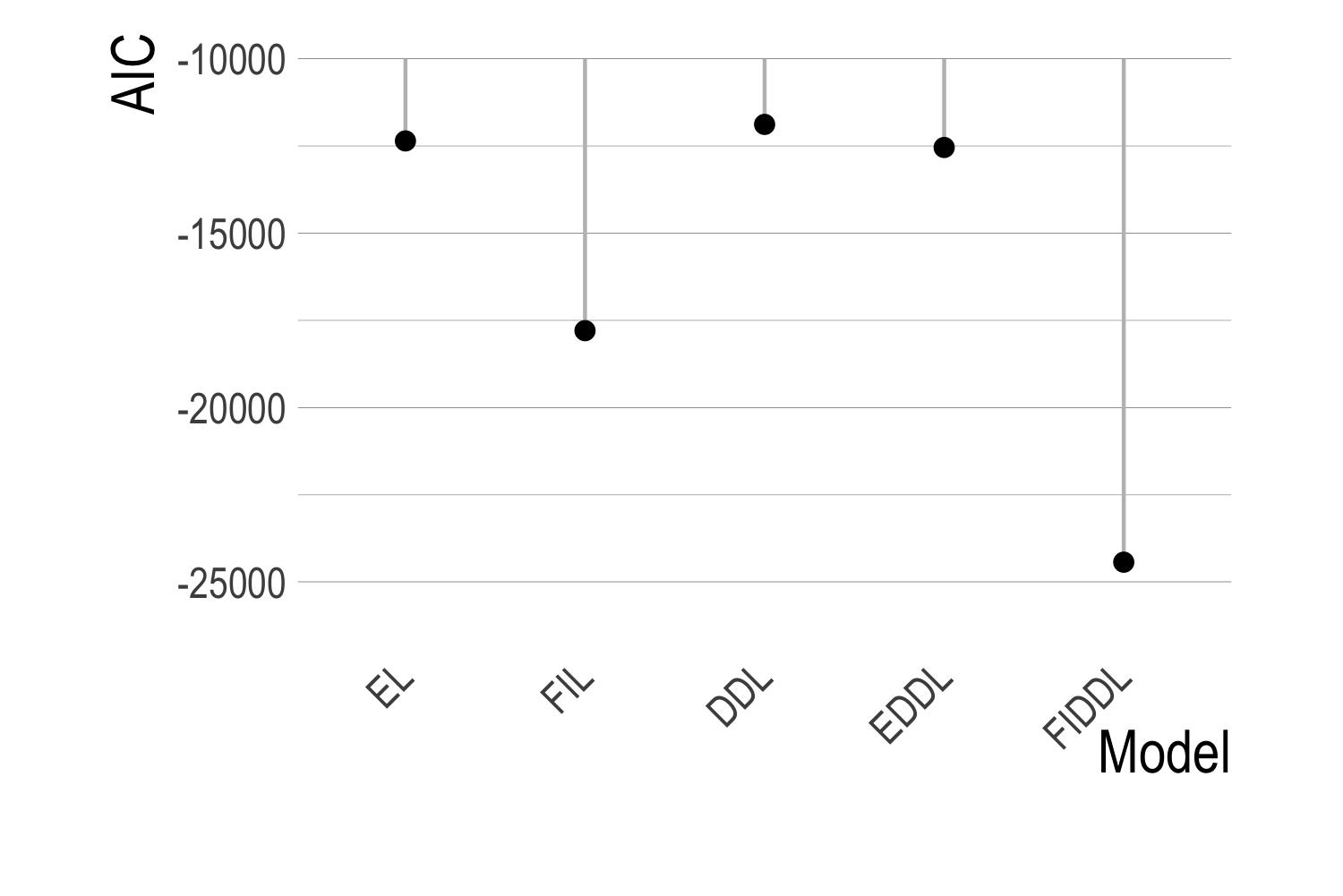}
        \caption{AIC for predicting RTs in the BLP}
    \end{subfigure}\hspace{0.5cm}
    \begin{subfigure}{.45\textwidth}
        \centering
        \includegraphics[width=\textwidth]{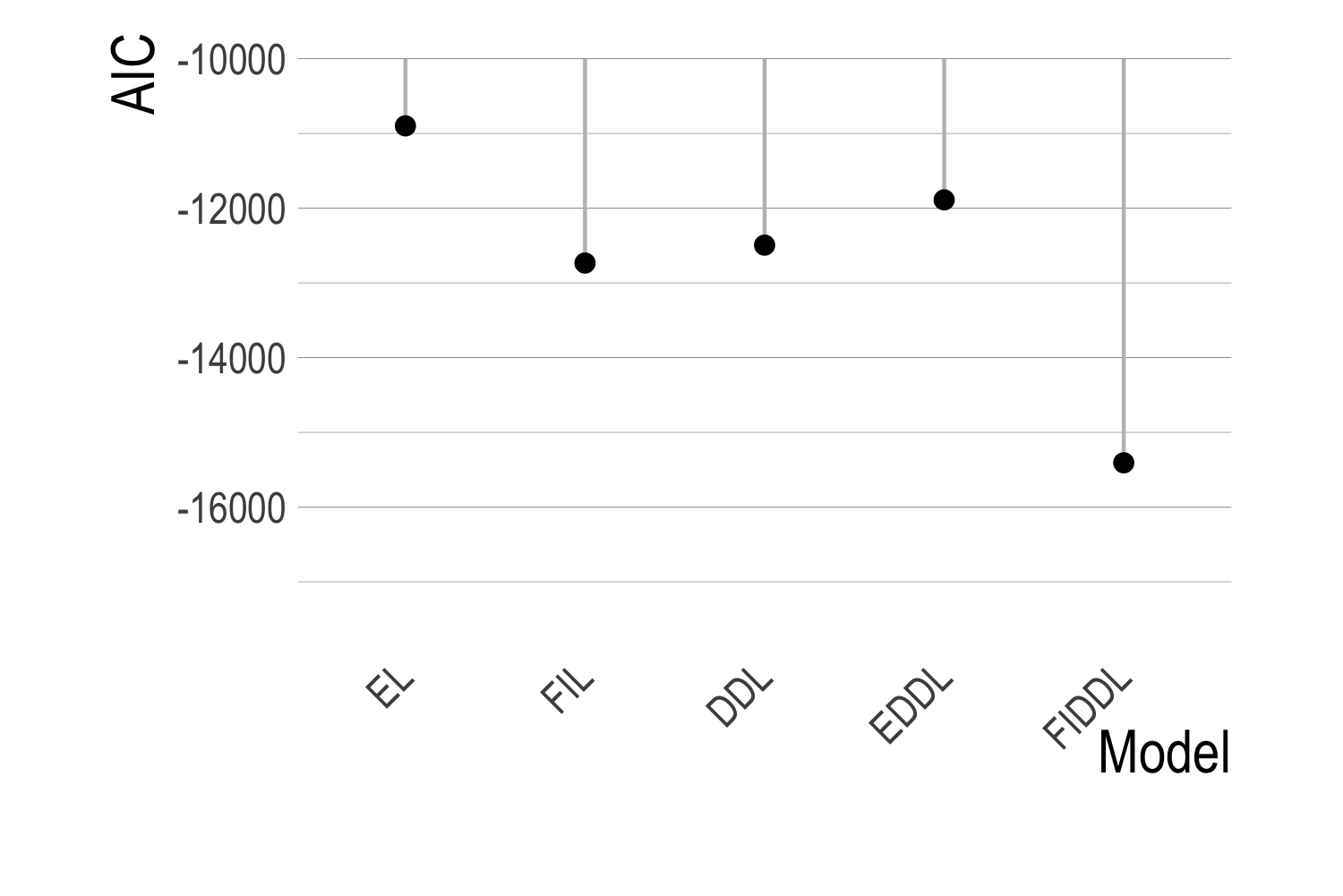}
        \caption{AIC for predicting RTs in the DLP}
    \end{subfigure}
    \caption{Target correlation taken from an EL, FIL, DDL, EDDL and FIDDL model predicting reaction times in the BLP and DLP. Figure adapted from \citet{heitmeier2024}.}
    \label{fig:dlp_blp_rts}
\end{figure}

\subsection{Predicting trial-to-trial reaction times}

Next, we turn to using DDL models for predicting more fine-grained data, namely lexical decision data at the per-participant trial-to-trial level. \citet{heitmeier2023trial} hypothesised that humans' lexical knowledge is continuously changing and that this continuous change can be modelled with the DLM. They simulated the lexical decision experiment reported in the BLP. For each participant they ran one simulation where they updated the LDL mappings in the DLM using the Widrow-Hoff \citep{Widrow1960} learning rule (dynamic simulation) and one where they did not update the mappings (static simulation). After applying one learning step of the Widrow-Hoff learning rule, the error between a predicted and target vector is slightly lower than before the update. This allows updating the mappings after each encountered stimulus in the lexical decision experiment in our simulations \citep[details in][]{heitmeier2023trial, heitmeier2024}. Measures from these simulations were then used to predict individual reaction times in the BLP. They found that for a majority of subjects, dynamic simulations provided a better model fit to reaction times, particularly for words (as opposed to nonwords).

Here, we asked whether we can replicate these findings with DDL. To this end we repeated the analysis presented in \citet{heitmeier2023trial} for the first 39 subjects in the BLP, replacing LDL mappings with DDL models and applying one backpropagation step (instead of the Widrow-Hoff learning rule) for updating the models after each trial. As in the linear simulations, each backpropagation step included calculating the loss of only this trial's predicted vector and using this loss to update the model's weights. Details on the simulation procedure, the measures used for predicting reaction times and the statistical modelling can be found in \citet{heitmeier2023trial} and \citet{heitmeier2024}.

\begin{figure}[!ht]
    \centering
    \includegraphics[width=.47\textwidth]{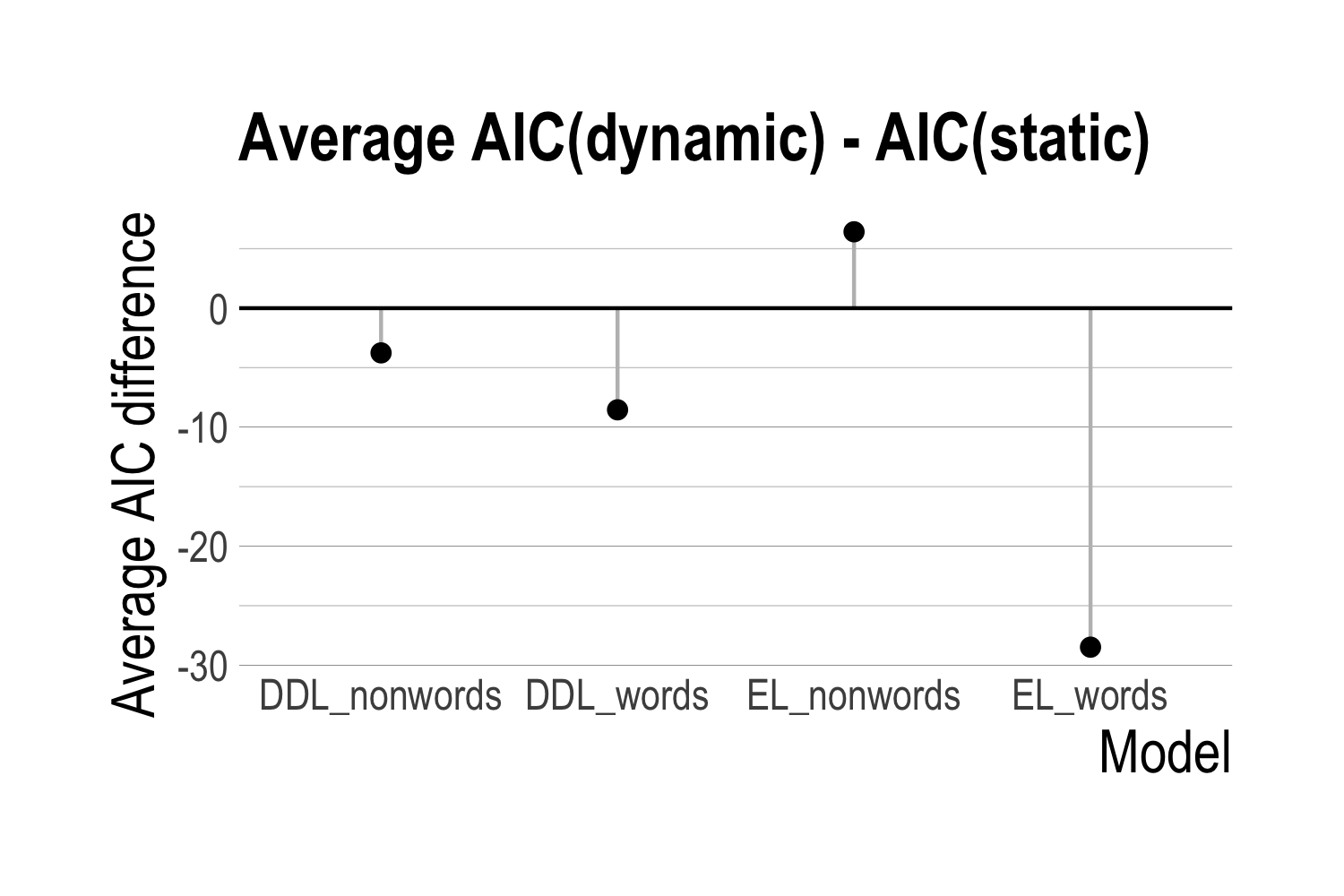}
    \includegraphics[width=.47\textwidth]{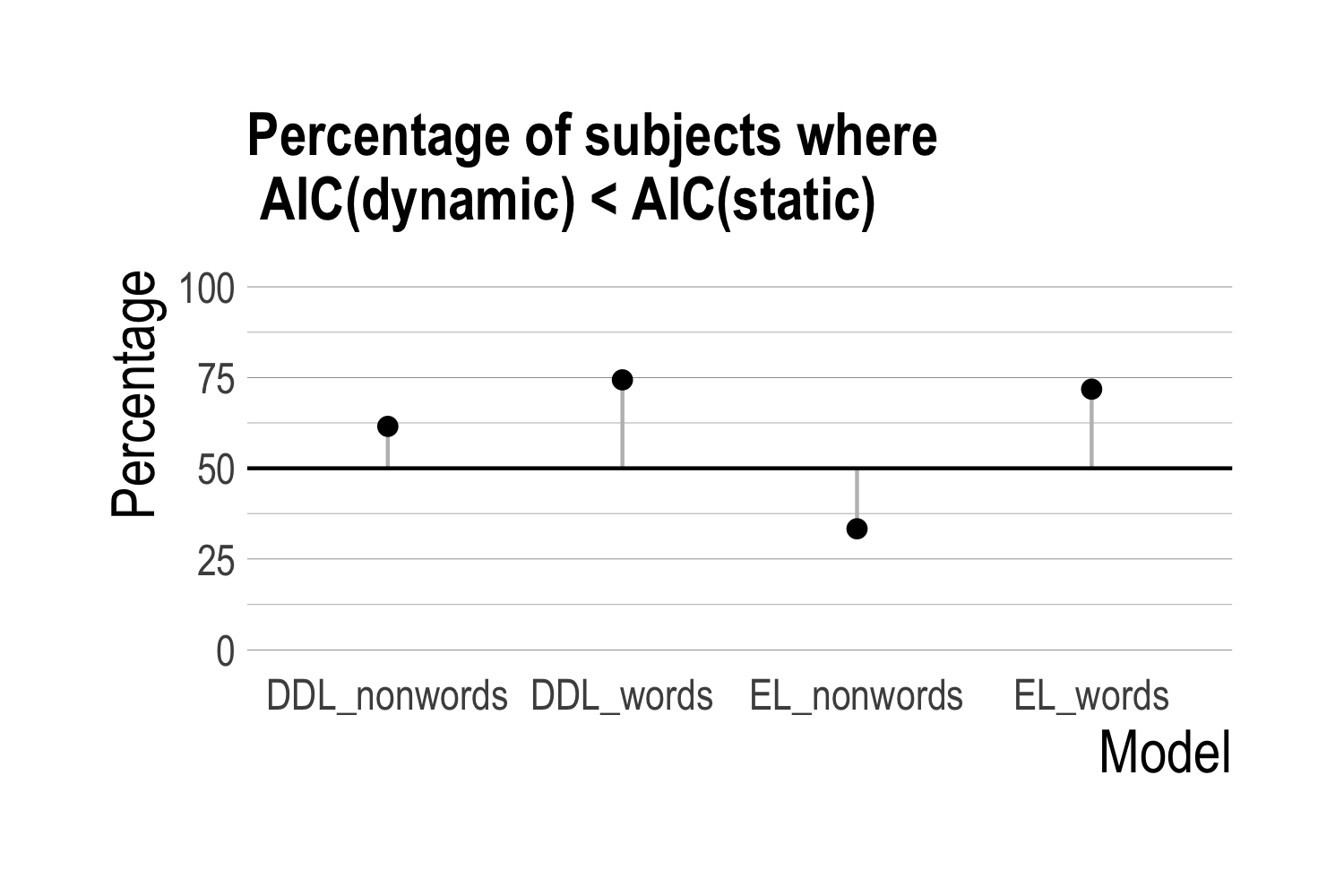}
    \caption{Comparison between static and dynamic simulations using DDL and LDL for predicting per-participant reaction times in the BLP. For words, the effect of dynamic simulations compared to static ones is larger for LDL, resulting in a bigger difference in AIC. Figure adapted from \citet{heitmeier2024}.}
    \label{fig:aic_trial_to_trial}
\end{figure}

We found that the AIC of all DDL-based models was worse than that of the LDL-based counterparts for all participants, for both words and nonwords and static/dynamic simulations (average AIC difference was -108.5/-128.4 AIC points for words for
static/dynamic simulations and -128.3/-118.1 AIC points for nonwords
for static/dynamic simulations). While for LDL we found that 72\% of dynamic simulations showed a better fit than static simulations for words (33\% for nonwords), for DDL 74\% of dynamic simulations outperformed static ones for words (62\% for nonwords). Figure~\ref{fig:aic_trial_to_trial} gives an overview over the average difference in AIC as well as the percentage of participants showing an advantage of trial-to-trial learning.

\begin{figure}[!ht]
    \centering
    \begin{subfigure}{.45\textwidth}
    \centering
        \includegraphics[width=\linewidth]{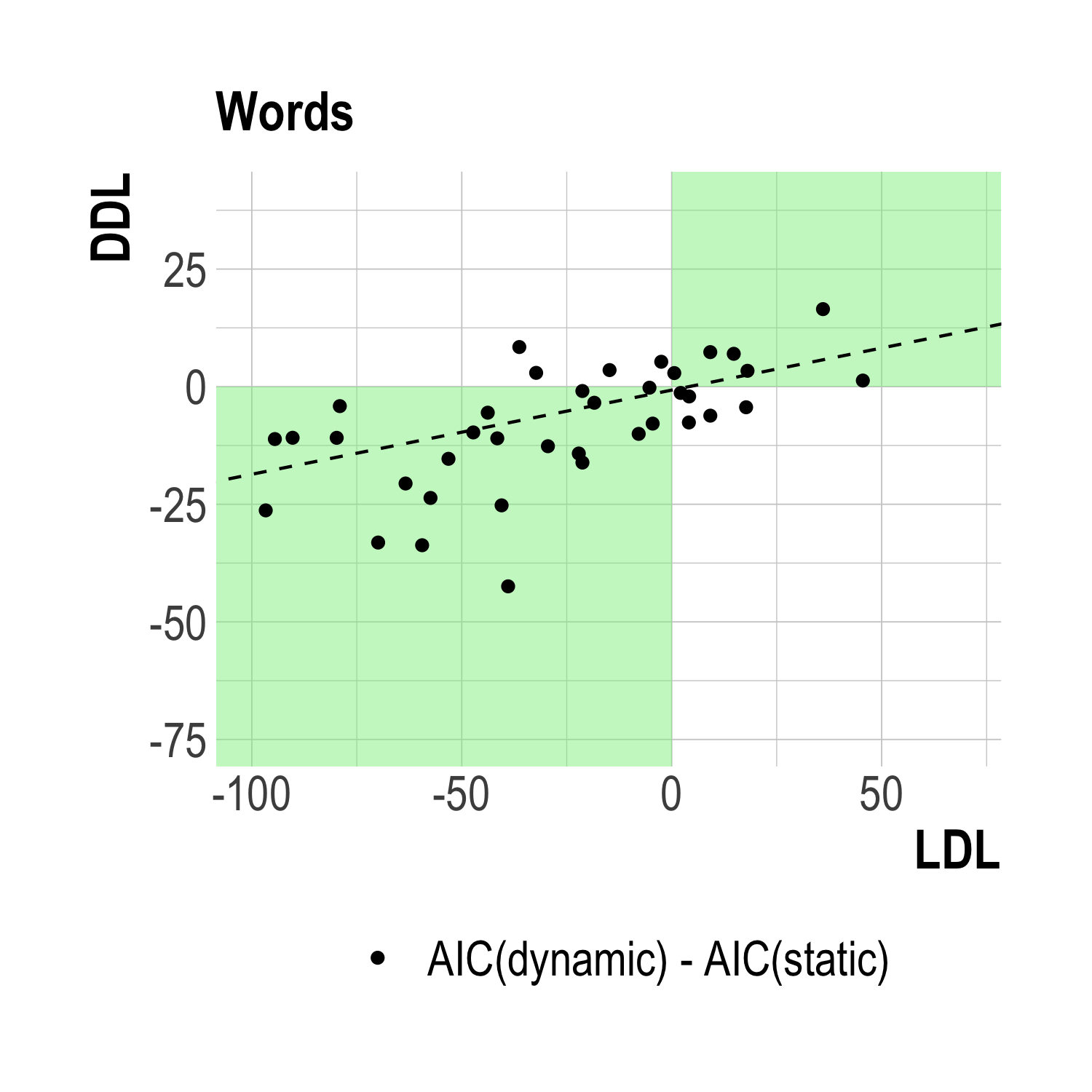}
        \caption{Words}
    \end{subfigure}
    \begin{subfigure}{.45\textwidth}
    \centering
        \includegraphics[width=\linewidth]{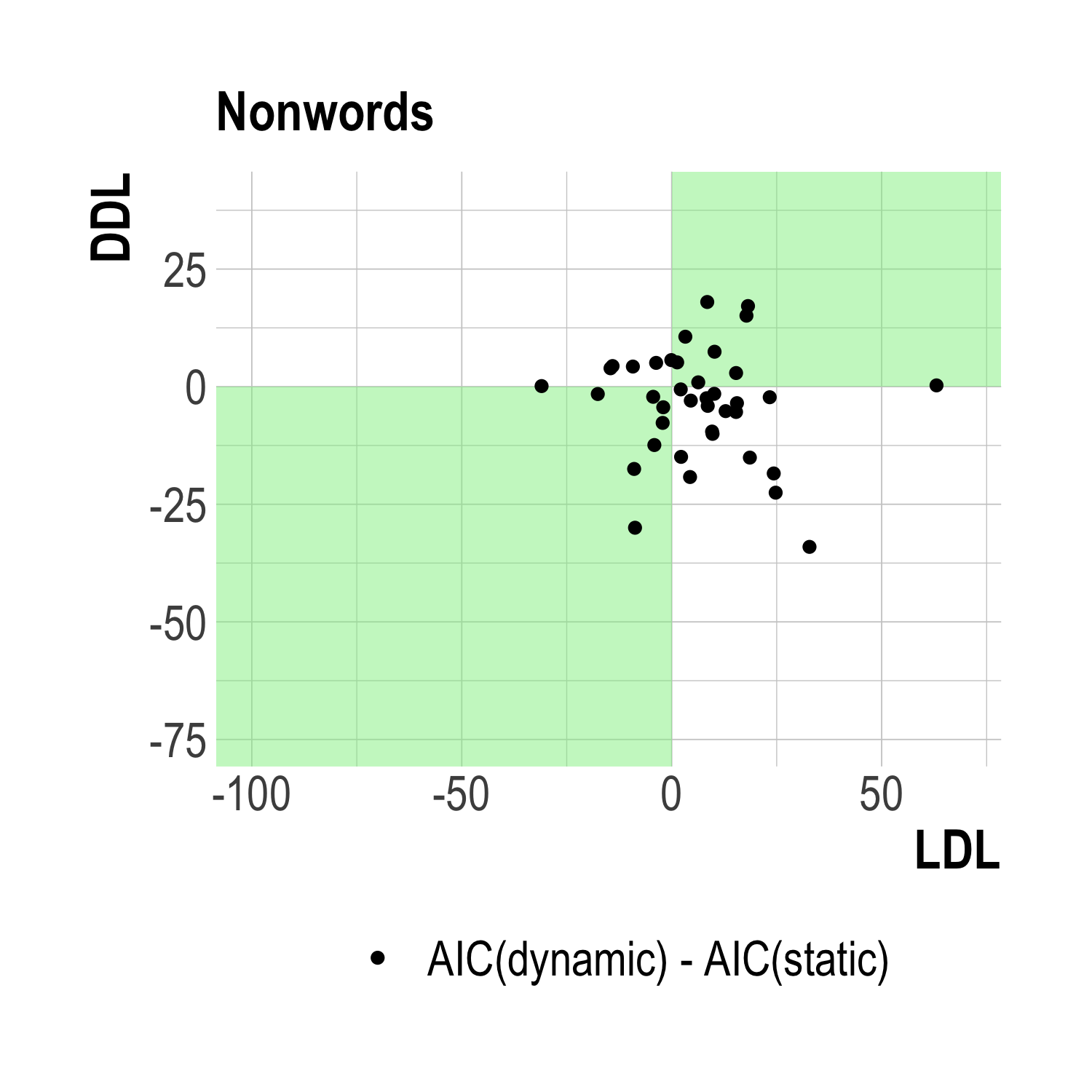}
        \caption{Nonwords}
    \end{subfigure}
    \caption{Comparison of AIC differences between dynamic and static simulations for DDL and LDL. Each dot represents one participant. The x-axis denotes the difference in AIC between dynamic and static simulations for the LDL models and the y-axis for the DDL models. Negative difference values indicate that the dynamic simulation outperforms the static simulation in terms of AIC, while positive difference values indicate the opposite. If a dot falls within a green shaded area, LDL and DDL agree on whether dynamic simulations provide a better model fit than static simulations for the respective participant. For words, the AIC differences are correlated ($r=.48, p=0.001$); for nonwords no correlation is present ($p=0.4715$). Figure adapted from \citet{heitmeier2024}.}
    \label{fig:aic_diff_corr}
\end{figure}

When plotting the difference between the AICs of static and dynamic simulations with DDL against that with LDL (Figure~\ref{fig:aic_diff_corr}), we found that they correlate for words, that is, DDL and LDL generally agreed on which subjects show an effect of trial-to-trial learning as modelled by the dynamic simulation. No such correlation was present for nonwords.

We conclude from this analysis that DDL is not able to model fine-grained trial-to-trial effects as well as LDL. For words, DDL shows an advantage of learning for a similar number of subjects as LDL but the difference between dynamic and static is smaller for DDL than for LDL. At least for words, DDL and LDL largely agree on which subjects show an effect of trial-to-trial learning.  A possible reason for the reduced effect of trial-to-trial learning in DDL is that a more accurate initial mapping may lead to less pronounced error feedback in the update step, thus each learning step having a smaller impact on subsequent trials. Moreover, the larger number of weights in the DDL networks might also mean that weight updates are distributed more evenly across weights and not centered as strongly on the weights of cues shared with subsequent words.

\section{Discussion and Conclusion}\label{sec:discussion}

In this paper we explored whether replacing the linear mappings (Linear Discriminative Learning, LDL) in the Discriminative Lexicon Model (DLM) with dense neural networks (termed Deep Discriminative Learning, DDL) improves the model's mapping accuracy. More crucially for cognitive modelling however, we also investigated whether such improved mapping accuracy also increases the model's performance at predicting behavioural data. We found that across four languages (English, Dutch, Taiwan Mandarin, Estonian) DDL always improved training accuracy and also accuracy on held-out data for English on comprehension, the dataset with the largest number of different lemmas. For production, DDL performed better than LDL across both training and test data for all four datasets. When predicting average lexical decision reaction times, we found that a frequency-informed variant of DDL (FIDDL) where the model is trained on each word according to its frequency outperformed all other methods (including frequency-informed LDL, FIL). However, when predicting fine-grained trial-to-trial lexical decision reaction times, DDL performs worse than LDL, even though the two generally agree as to which subjects show a better model fit when based on simulations where the DLM is dynamically updated after each trial.

It is worthwhile to focus on a few aspects of these results in more detail. First, we find a notable difference across datasets: In comprehension, DDL only reliably improves mapping accuracy over that of LDL for English, but not for Mandarin and Estonian. This may to some extent be explainable by the respective data sizes: the English dataset is larger (28,000 word forms) than the Estonian and Mandarin ones (6,000 and 3,000). However, the nature and diversity of the datasets may also play role. The English and Dutch datasets have a much larger number of different lemmas with few inflected word forms per lemma, while the Estonian dataset only has 230 lemmas with a large number of (relatively regular) inflected forms. For Mandarin, tone information contributes to a much larger number of cues (9,644, compare this to the much larger English dataset, which only has 7,030 cues), and as we saw in Section~\ref{sec:internal}, fewer shared cues increase the performance of LDL models in comprehension. This suggests that the more semantically diverse a dataset, the more it profits from modelling with DDL. In production, on the other hand, all four models profit from DDL. In \citet{heitmeier2024} we found that this can be accounted for to a significant extent by the loss function: we used the binary cross-entropy loss to train a linear model with the same number of trainable parameters as LDL but with an added sigmoid function and found that it achieved almost the same accuracy as the deep learning models. However, the advantage of DDL over LDL in the present setup may also be due to the representations used. \citet{chuang2024word} found that LDL performs surprisingly well at mapping contextualised embeddings to pitch contours, and preliminary results suggest that in this setup, DDL does not improve over LDL.\footnote{However, mapping accuracy from form (low-dimensional) to embeddings (high-dimensional) did improve compared to linear mappings, which is perhaps unsurprising given the difference in dimensionality.}
This indicates that when both input and output are sufficiently discriminable, LDL suffices.

Second, notice that while DDL models outperformed LDL models in terms of pure mapping accuracy at least for the English dataset, they do not necessarily outperform LDL at predicting reaction times: DDL outperforms LDL at predicting average reaction times in the Dutch Lexicon Project (DLP), but not in the British Lexicon Project (BLP). The same is true for modelling lexical decision reaction times at the individual subject-level: even leaving aside trial-to-trial learning, the DDL models never perform better than the LDL models for static simulations. This highlights that pure mapping accuracy is not enough for predicting reaction times. However, it also leads to a further point, namely that modern machine learning and artificial intelligence may in some aspects be too good: human speakers make a lot of mistakes and are not particularly accurate. For example, participants in the DLP agree with themselves at chance level for low frequency words (below 10 per million) when tested on the same words twice \citep{diependaele2012noisy}. In the study of \citet{ramscar2013learning}, young adults (mean age 21.1 years) performed at chance level for low frequency words in an English lexical decision task \citep{balota1999item}. Moreover, the question arises of whether fully optimized performance in the DLM would be in any way helpful for predicting reaction times. Dense neural networks can in principle learn any continuous function to an arbitrary point of accuracy when given enough parameters \citep{hornik1989multilayer}. In the case of the DLM this would mean that all predicted semantic and form vectors would be entirely accurate, thus losing any information we can gain from the DLM above and beyond simply using binary form vectors and pre-trained embeddings. The DLM's imprecisions and mistakes offer valuable insights into possible processing bottlenecks.

Third, examining the words that DDL maps more accurately than LDL is illuminating for linguistic analysis. We found that LDL outperforms DDL for words which contain very unique trigrams (e.g. \textit{\#px}, \textit{\#oz}), while DDL outperforms LDL when words contain trigrams which express a range of different functions such as \textit{-er}. The re-use of building blocks is ubiquitous in language. While humans are able to deal with this ``bricolage''  relatively effortlessly, it is nevertheless a processing bottleneck which LDL struggles with. In this way, the comparison between LDL and DDL shows which components of single word processing may be particularly challenging. This finding could also have implications for theories of morphological priming \citep[e.g.][]{taft1975lexical, rastle2008morphological}: our results suggest that LDL may place pseudomorphological words (e.g. \textit{barriers}, \textit{choler}) in the semantic neighbourhood of morphologically complex words (e.g. \textit{livelier}, \textit{chunkier}). This could explain why morphologically transparent and morphologically opaque words are primed in a similar way \citep[e.g.][]{creemers2020opacity} even without morphological decomposition \citep[for linear DLM models capturing morphological priming effects, see][]{baayen2020modeling,chuang2021vector}. Again, however, this result may, at least in part, be due to the representations used. An examination of the final segment of words ending in \textit{-er} in the Buckeye corpus \citep{pitt2007buckeye} reveals that they have more varied pronunciations in nouns than in adjectives. These results dovetail well with the hypothesis that the nouns, which have much more varied meanings, have more differentiated realizations of the exponent --- a distinction that is neutralized in the orthography. This in turn suggests that deep learning is required for \textit{-er} words in English because the representations are impoverished, which opens the possibility that a linear mapping based on actual articulation might suffice.

Fourth, our results highlight the importance of the data distribution the models are trained on for cognitive modelling. Using a training/validation/test split is not a plausible mechanism for training cognitive models: after all, we usually encounter high frequency (seen) words, with the occasional low frequency (possibly unseen) word interspersed in-between. FIDDL (frequency-informed DDL) therefore has a much more plausible learning setup: it is trained on the full token distribution. By calculating the model's accuracy across all words in the dataset we get a more realistic picture of how the model would fare with a real-world distribution of tokens \citep[see also][for a discussion of token-based accuracy calculations]{Heitmeier:Chuang:Axen:Baayen:2022}. Surprisingly, this accuracy is very high (97\% and 87\% for Dutch and English respectively) even though the model saw 40\% of the words in the English and 9\% of the words in the Dutch dataset only 10 times or fewer. Such a setup also quells concerns about the performance of computational models on held-out data \citep{pinker2002past}. Humans are able to generalise to unseen data, but in reality, word distributions are skewed towards high frequency words. Evidently, both humans and FIDDL perform excellently in such a setting. Nonetheless, one might argue that particularly in languages with rich morphology such as Finnish, speakers may still need to generalise to low frequency unseen word forms regularly.  As shown by \citet{NikolaevJoensuu} for a set of some 50,000 Finnish inflected nouns, EL can already perform very well on both training data and held-out low-frequency words. To address this question with further precision,  a natural next step is to evaluate performance of FIDDL on these and similar data sets.
In any case, these results once again highlight the importance of training models of lexical processing in a frequency-informed way when one is interested in a usage-based analysis of the data. The frequency-informed FIDDL and FIL models by far outperform their non-frequency informed counterparts when predicting average reaction times in the BLP and DLP.

To summarise, researchers interested in modelling a language system or behavioural data with the DLM may well profit from using DDL when they are working with a particularly large, diverse dataset which LDL struggles with and/or when they require a higher mapping accuracy for their modelling task. However, they should be aware that using DDL may not necessarily improve the model's ability to predict behavioural data (unless trained in a frequency-informed way) and that DDL is costly: it requires experimenting with a number of hyperparameters as well as significantly more computational resources and training times.

Where does this leave us with regards to our initial question about deep learning's role in understanding the learning problem which speakers face? Our comparison between LDL and DDL in the DLM lead us to a few conclusions. While deep learning models are generally more accurate, they are not a ``tried-and-true'' method for increasing cognitive modelling performance: pure accuracy improvements do not necessarily afford better performance at predicting behavioural data and may also not improve our understanding of the underlying processes. However, investigating cases where deep learning models clearly outperform linear models can nevertheless point to processing bottlenecks in a purely linear system which can be solved at the expense of additional parameters in deep learning models.

\section*{Acknowledgements}
Funded by the Deutsche Forschungsgemeinschaft (DFG, German Research Foundation) under Germany’s Excellence Strategy – EXC number 2064/1 – Project number 3907276455, and by the European Research Council, project WIDE-742545. The authors thank Yu-Ying Chuang, Tino Sering and Louis ten Bosch for helpful discussions.

\bibliographystyle{apalike}
\bibliography{bib_clean}

\end{document}